\documentclass[conference]{IEEEtran}
\IEEEoverridecommandlockouts

\usepackage{cite}
\usepackage{amsmath,amssymb,amsfonts}
\usepackage{algorithmic}
\usepackage{graphicx}
\usepackage{textcomp}
\usepackage{xcolor}
\usepackage{hyperref}
\usepackage{multirow}
\usepackage{adjustbox}
\usepackage{booktabs}
\def\BibTeX{{\rm B\kern-.05em{\sc i\kern-.025em b}\kern-.08em
    T\kern-.1667em\lower.7ex\hbox{E}\kern-.125emX}}
\begin{document}

\title{Generative Models for Helmholtz Equation Solutions: A Dataset of Acoustic Materials

\thanks{$^*$Equal contribution.}
}

\author{\IEEEauthorblockN{Riccardo Fosco Gramaccioni\textsuperscript{1}$^{*}$, Christian Marinoni\textsuperscript{1}$^{*}$, Fabrizio Frezza\textsuperscript{1},\\ Aurelio Uncini\textsuperscript{1}, and Danilo Comminiello\textsuperscript{1}} \\
\IEEEauthorblockA{\textsuperscript{1}Department of Information Engineering, Electronics and Telecommunications (DIET),\\Sapienza University of Rome, Rome. Italy,\\ \{riccardofosco.gramaccioni; christian.marinoni\}@uniroma1.it}
}

\maketitle

\begin{abstract}
Accurate simulation of wave propagation in complex acoustic materials is crucial for applications in sound design, noise control, and material engineering. Traditional numerical solvers, such as finite element methods, are computationally expensive, especially when dealing with large-scale or real-time scenarios. In this work, we introduce a dataset of 31,000 acoustic materials, named HA30K, designed and simulated solving the Helmholtz equations.
For each material, we provide the geometric configuration and the corresponding pressure field solution, enabling data-driven approaches to learn Helmholtz equation solutions.
As a baseline, we explore a deep learning approach based on Stable Diffusion with ControlNet, a state-of-the-art model for image generation. 
Unlike classical solvers, our approach leverages GPU parallelization to process multiple simulations simultaneously, drastically reducing computation time. By representing solutions as images, we bypass the need for complex simulation software and explicit equation-solving. Additionally, the number of diffusion steps can be adjusted at inference time, balancing speed and quality. We aim to demonstrate that deep learning-based methods are particularly useful in early-stage research, where rapid exploration is more critical than absolute accuracy.

\end{abstract}


\section{Introduction}

The Helmholtz equation provides a fundamental mathematical framework for modeling steady-state wave behavior in complex acoustic materials, enabling the prediction of sound pressure fields under different configurations. However, solving the Helmholtz equation numerically is computationally expensive, particularly for high-resolution simulations or real-time applications \cite{oberai2000numerical}.

\begin{figure}[!t]
    \centering
    \includegraphics[width=0.9\linewidth]{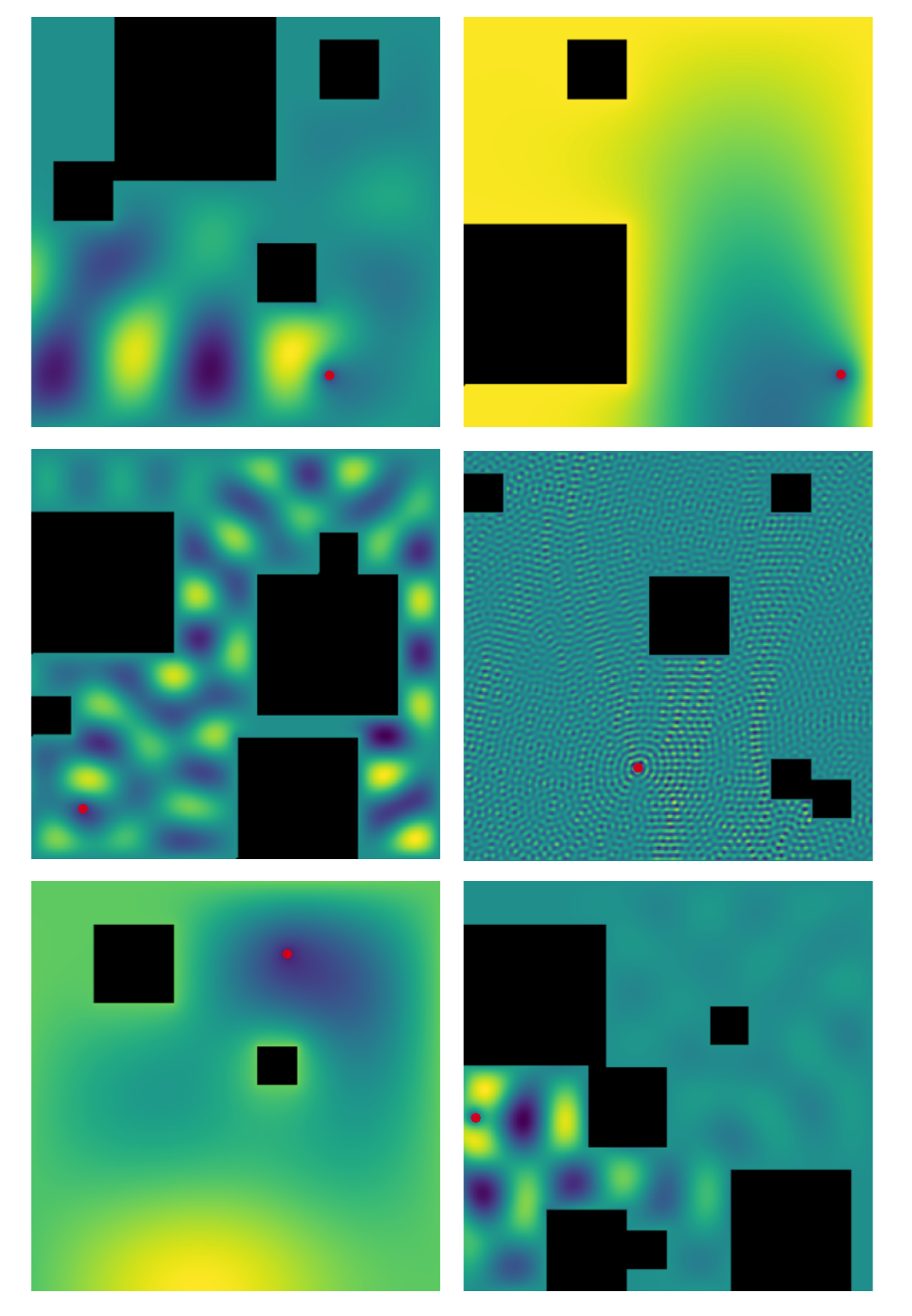}
    \caption{Samples from the HA30K dataset. The red dot represents the sound source location. Black squares represent the obstacles.}
    \label{fig:examples}
\end{figure}

Traditional numerical methods, such as the finite element method (FEM) and finite difference method (FDM), provide accurate solutions but require significant computational resources, making them impractical for large-scale or interactive applications. Recent advancements in deep learning have shown promise in accelerating acoustic \cite{gramaccioni2024l3das23, marinoni2023overview} and physics-based simulations by learning data-driven mappings between material configurations and their corresponding physical responses. Such approaches have been explored in fluid dynamics \cite{tompson2017accelerating}, heat transfer \cite{cai2021physics}, and electromagnetism \cite{raissi2019physics}, but their application to acoustic wave propagation remains relatively underexplored.

In this work, we introduce the \textit{Helmholtz Acoustic 30K} (HA30K) dataset, designed to bridge the gap between computational acoustics and deep learning. The dataset consists of 31,000 2D acoustic material configurations generated using FreeFEM, a finite element solver for partial differential equations. Each sample includes a square domain with one to six square obstacles, each made of different materials such as rubber, wood, or metal. The acoustic source is modeled as a Gaussian source, randomly placed within the non-occupied regions of the domain. For each configuration, we provide both the material distribution image and the corresponding pressure field $p$ solution obtained by solving the Helmholtz equation. Some examples are depicted in Figure~\ref{fig:examples}. The dataset can be downloaded from Zenodo\footnote{https://zenodo.org/records/15683385}.

The Helmholtz equation, which governs the steady-state behavior of acoustic waves, depends on key parameters such as the wave number $k$, which encapsulates frequency and material properties, and appropriate boundary conditions to account for wave reflections and absorptions. In our dataset, we enforce conditions that simulate real-world wave interactions, including absorbing boundaries to prevent artificial reflections and Dirichlet conditions on obstacles to model rigid surfaces. The resulting pressure fields provide a rich representation of wave propagation patterns across different material configurations.

To evaluate the potential of deep learning for acoustic simulation, we use as a baseline Stable Diffusion with ControlNet, since diffusion models have demonstrated excellent capabilities in generating high quality images \cite{sigillo2024ship}. We trained the model to generate accurate pressure field predictions from material configuration images, offering a fast and data-driven alternative to conventional numerical solvers.

This paper aims to provide a benchmark for future research in applying deep learning to acoustic material simulation. By releasing our dataset and baseline models, we hope to inspire further exploration of deep learning-driven solutions for computational acoustics, enabling faster and more scalable simulations for real-world applications.

Our contribution can be summarized as follows:
\begin{itemize}

\item We introduce a novel dataset of 31,000 2D acoustic material configurations, including pressure field solutions obtained using FreeFEM.

\item We test our dataset on a baseline model derived from Stable Diffusion with ControlNet to predict Helmholtz equation solutions from material configurations.

\item We perform an evaluation based on some of the most common objective metrics for image quality estimation and a comparative analysis between traditional numerical solvers and deep learning-based approaches, highlighting their advantages and limitations for acoustic wave simulation.
\end{itemize}

The rest of the paper is organized as follows. Section~\ref{sec:works} presents the related works, Section~\ref{sec:dataset} the proposed dataset structure, while in Section~\ref{sec:exp} we discuss the baseline model and we validate the obtained experimental results. Finally, conclusions are drawn in Section~\ref{sec:con}.

\section{Related Works}
\label{sec:works}

Deep learning has emerged as a powerful tool for advancing the design and analysis of complex materials, including acoustic materials. Recent studies demonstrate the potential of generative models and deep learning-based inverse design techniques in discovering novel material properties and structures. Diffusion models, in particular, have gained popularity as physics-informed approaches capable of generating physically plausible designs while maintaining computational efficiency \cite{bastek2025physicsinformed, Bastek2023InversedesignON}. These methods have been successfully applied to inverse-design tasks for nonlinear mechanical metamaterials and multi-modal resonant structures \cite{Dedoncker2023GenerativeID}.

For acoustic materials, deep learning has proven effective in optimizing surfaces for sound absorption. Different studies have used machine learning techniques to design materials with high degrees of freedom \cite{Donda2021UltrathinAA, Donda2022DeepLA}, enabling the automated discovery of novel acoustic structures. The application of inverse design has also extended to phononic crystals and multishape metamaterials, allowing for tailored wave propagation characteristics \cite{Dykstra2023InverseDO, Ha2023RapidID}.

In wider terms,  advancements in data-driven methods have demonstrated the feasibility of using deep learning to optimize materials for specific mechanical, optical, and electromagnetic properties \cite{Liu2021MachineLP, Liu2022AllangleRN, gramaccioni2024inverse, Muhammad2022MachineLA}. Several works have explored deep learning as a tool for accelerating the discovery of new material configurations, facilitating rapid generation and evaluation of candidate structures \cite{Sarkar2023PhysicsInformedML, Schneuing2022StructurebasedDD}. Notably, generative models have been employed for mechanistic-based learning and on-demand inverse design of metamaterials, highlighting the increasing role of deep learning in material science \cite{Wang2020DeepGM, Wang2023OndemandID}.

Building on this foundation, our work presents a dataset of simulated acoustic materials designed using the Helmholtz equation, providing a benchmark for evaluating deep learning-based methods for acoustic simulation. Unlike previous studies focusing on specific material categories or optimization tasks, our dataset aims to facilitate the development of generalizable models capable of predicting acoustic wave behavior in diverse material configurations.

\section{HA30K Dataset}
\label{sec:dataset}

In this section, we introduce the HA30K dataset—a collection of 31,000 numerical samples derived from solutions of the Helmholtz equation. These samples capture a wide range of acoustic scenarios, making the dataset a valuable resource for research in wave propagation, numerical simulation, and machine learning applications in acoustics. All numerical solutions were computed using FreeFEM, a open-source high-level multiphysics finite element solver.

\subsection{Theoretical Background}

The propagation of acoustic waves is modeled by the Helmholtz equation. In its strong form, the equation is given by

\begin{equation}
    \Delta p + k^2 p = f \quad \text{in } \Omega,
\end{equation}

where \( p \) denotes the pressure field, \( k \) is the wave number, and \( f \) is a given source term. To ensure a well-posed problem, we impose boundary conditions, such as the Sommerfeld radiation condition on the boundary of the domain \(\Gamma_1\), representing the boundary of the domain \(\Omega\):

\begin{equation}
    \frac{\partial p}{\partial n} - i k p = 0 \quad \text{on } \Gamma_1,
\end{equation}

and a Robin boundary condition on obstacle surfaces \(\Gamma_{\text{obstacles}}\):

\begin{equation}
    \frac{\partial p}{\partial n} + i Z p = 0 \quad \text{on } \Gamma_{\text{obstacles}},
\end{equation}
where $Z$ is the impedence parameter.
\begin{figure}[!t]
    \centering
    \includegraphics[width=\linewidth]{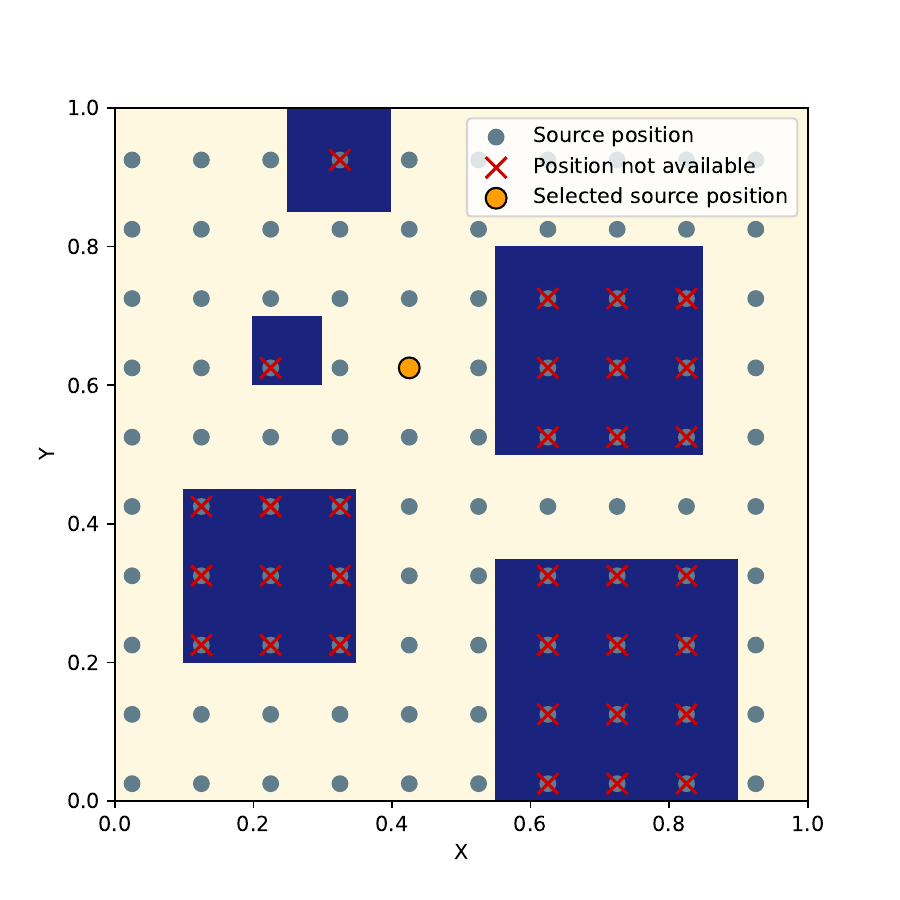}
    \caption{An example showing the possible locations of the sound source given the position of obstacles (in blue). }
    \label{fig:grid}
\end{figure}
For numerical simulations, we derive the weak formulation by multiplying the strong form by a test function \(q\), integrating over the domain \(\Omega\), and applying integration by parts. The resulting weak formulation is:
\begin{align}
    &\underbrace{\int_{\Omega} \left(\nabla p \cdot \nabla q - k^2 p\,q \right) dx}_{\text{Weak formulation of the Helmholtz eq.}} \notag
    \quad + \underbrace{\int_{\Omega} f\,q\,dx}_{\text{Gaussian source}} \notag \\
    &\quad - \underbrace{\int_{\Gamma_{1}} i\,k\, p\,q\,ds}_{\text{Sommerfeld boundary condition}} \notag
    \quad + \underbrace{\int_{\Gamma_{\text{obstacles}}} i\,Z\, p\,q\,ds}_{\text{Robin boundary condition}}= 0.
\end{align}


The Sommerfeld condition is imposed to simulate free radiation (i.e., to avoid reflections at the domain boundaries), while the Robin condition on obstacles models impedance effects. By varying the parameter \(Z\), we can simulate different obstacle materials (i.e., \(Z=150\) for foam, \(Z=600\) for rubber, \(Z=1500\) for wood, \(Z=1\times10^6\) for metal).

\begin{figure*}[!t]
    \centering
    \includegraphics[width=\linewidth]{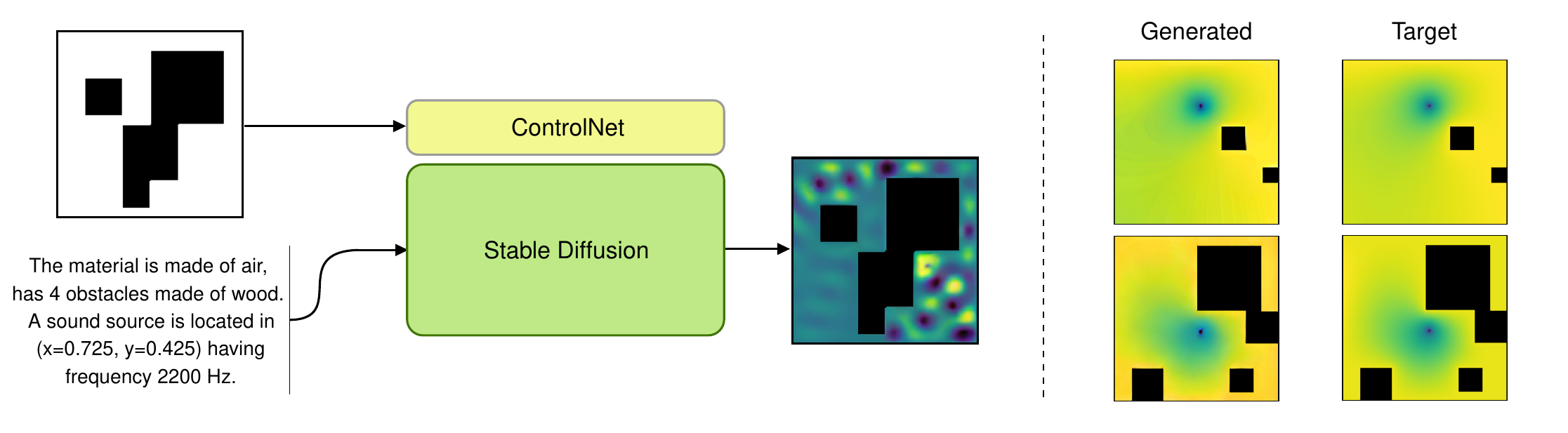}
    \caption{\textbf{Baseline architecture}: The baseline architecture takes the topological structure of the domain along with a textual description of the acoustic properties and outputs an estimate of the pressure field. The target is the pressure field obtained by solving the Helmholtz equation.}
    \label{fig:architecture-results}
\end{figure*}

\subsection{Domain definition}

The generation of each sample in the HA30K dataset involves the definition of the domain and its characteristics.
The domain is a unit square discretized with a 256×256 mesh. Within this domain, the speed of sound \(c\) is set according to the material properties of the medium, influencing the wave propagation characteristics. Table~\ref{tab:table materials} summarizes the values of $c$ given the different materials. Similarly, the frequency of the sound source can have a value between 100 and 4000 Hz (with increments of 100 Hz). Additionally, obstacles are modeled as squares and are randomly placed in non-overlapping configurations. Their dimensions are sampled from the interval $[0.1, 0.4]$ (with a step of 0.05) and their number varies from 1 to 6. Larger obstacles are assigned a higher probability when the total number of obstacles is lower, where on average 32\% of the domain surface occupied by obstacles. This flexible domain setup enables the simulation of a wide range of acoustic scenarios.
\begin{table}[]
    \centering
    \caption{Speeds of sound $c$ given different domain materials.}
    \resizebox{0.68\linewidth}{!}{
    \begin{tabular}{cc}
    \toprule
        \textbf{Material} & \textbf{Speed of sound (m/s)} \\
    \midrule
        Rubber & $60$ \\
        Air at 20°C & $343$ \\
        Lead & $1210$ \\
        Gold & $3240$ \\
        Glass & $4540$ \\
        Aluminum & $6320$ \\
    \bottomrule
    
    \end{tabular}}
    \label{tab:table materials}
\end{table}

\subsection{Implementation with FreeFEM}

FreeFEM was employed to solve the weak formulation of the Helmholtz equation using the finite element method. Its flexibility allowed us to incorporate complex boundary conditions and heterogeneous material properties, thereby enabling the generation of a large number of simulations under varying physical and geometrical configurations.
The solution \(p\) to the Helmholtz equation is complex-valued. For each simulation, the results are stored as a set of values \([X,\ Y,\ \text{Re}(p),\ \text{Im}(p)]\).
From the distribution of obstacles we derive the source images of the dataset, which represent with black pixels the obstacles and with white pixels the free portion of the domain. For the target images, we convert the real component of the pressure field (\text{Re}($p$)) into RGB images by applying the viridis colormap from matplotlib. The resulting dataset consists of 31,000 pairs of source and target images, each with dimensions of 256×256 pixels.



\section{Experimental Results}
\label{sec:exp}

\subsection{Baseline}
As a baseline for evaluating the effectiveness of deep learning methods in predicting the Helmholtz equation solutions, we employ Stable Diffusion with ControlNet \cite{zhang2023adding}. We use official repository, available on GitHub \footnote{\url{https://github.com/lllyasviel/ControlNet}} and the provided checkpoints. The model is trained on our dataset, which is split into training (80\%), validation (15\%), and test (5\%) sets.  

The input to ControlNet consists of the image representing the material domain, which includes the surface and obstacles, providing spatial information about the acoustic environment. Additionally, global parameters such as frequency, material of the surface, material of the obstacles, number of obstacles, frequency, and position of the sound source are encoded as a single text string having structure: “The material is made of \textit{\{material\_domain\}}, has \textit{\{num\_obstacles\}} obstacles made of \textit{\{material\_obstacles\}}. A sound source is located in (x=\textit{\{$s_x$\}}, y=\textit{\{$s_y$\}}) having frequency \textit{\{source\_freq\}} Hz". The text string is used as cross-attention conditioning for the model.
The target is the 256x256 pixel image derived by the real components of the pressure values $\text{Re}(p)$, as resulted from the simulations.

During training, the Stable Diffusion weights remain frozen, and we train only the ControlNet parameters to adapt the diffusion process to our specific dataset. This setup ensures that the model learns to generate pressure field solutions while leveraging the powerful generative capabilities of diffusion models. 
A block diagram of the baseline architecture is proposed in Figure~\ref{fig:architecture-results}.

The model is trained on a single 48 GB Nvidia RTX A6000. During training, we used a batch size of 4 with a learning rate kept fixed at $1e-5$ for a total of 100k diffusion steps, where each step is defined by 1000 timesteps. We use the Stable Diffusion 2.1\footnote{\url{https://huggingface.co/stabilityai/stable-diffusion-2-1-base/tree/main}} checkpoint and we keep the model frozen, training only the ControlNet parameters. Consequently, we use the original values for all other hyperparameters of the model. 
 
\subsection{Objective Evaluation}
To demonstrate the performance of the baseline used to test our dataset in predicting the solutions of the Helmholtz equation, we employ the following objective evaluation metrics:  
\begin{itemize}
\item Mean Squared Error (\textbf{MSE}): this metric measures the average squared difference between the predicted and ground truth pressure field images. A lower MSE indicates a higher accuracy in the reconstruction of the pressure distribution.  
\item Fréchet Inception Distance (\textbf{FID}): FID compares the statistical distribution of generated and real images in a feature space extracted from a pre-trained neural network. Lower FID values indicate that the generated pressure fields are perceptually closer to the ground truth.  
\item Structural Similarity Index Measure (\textbf{SSIM}): SSIM assesses the perceptual similarity between predicted and reference images by considering luminance, contrast, and structural information. Higher SSIM values indicate better structural preservation in the generated pressure fields. 
\end{itemize}

These metrics provide a comprehensive evaluation of both numerical accuracy and perceptual quality for the acoustic material simulations.

\begin{figure}[!t]
    \centering
    \includegraphics[width=\linewidth]{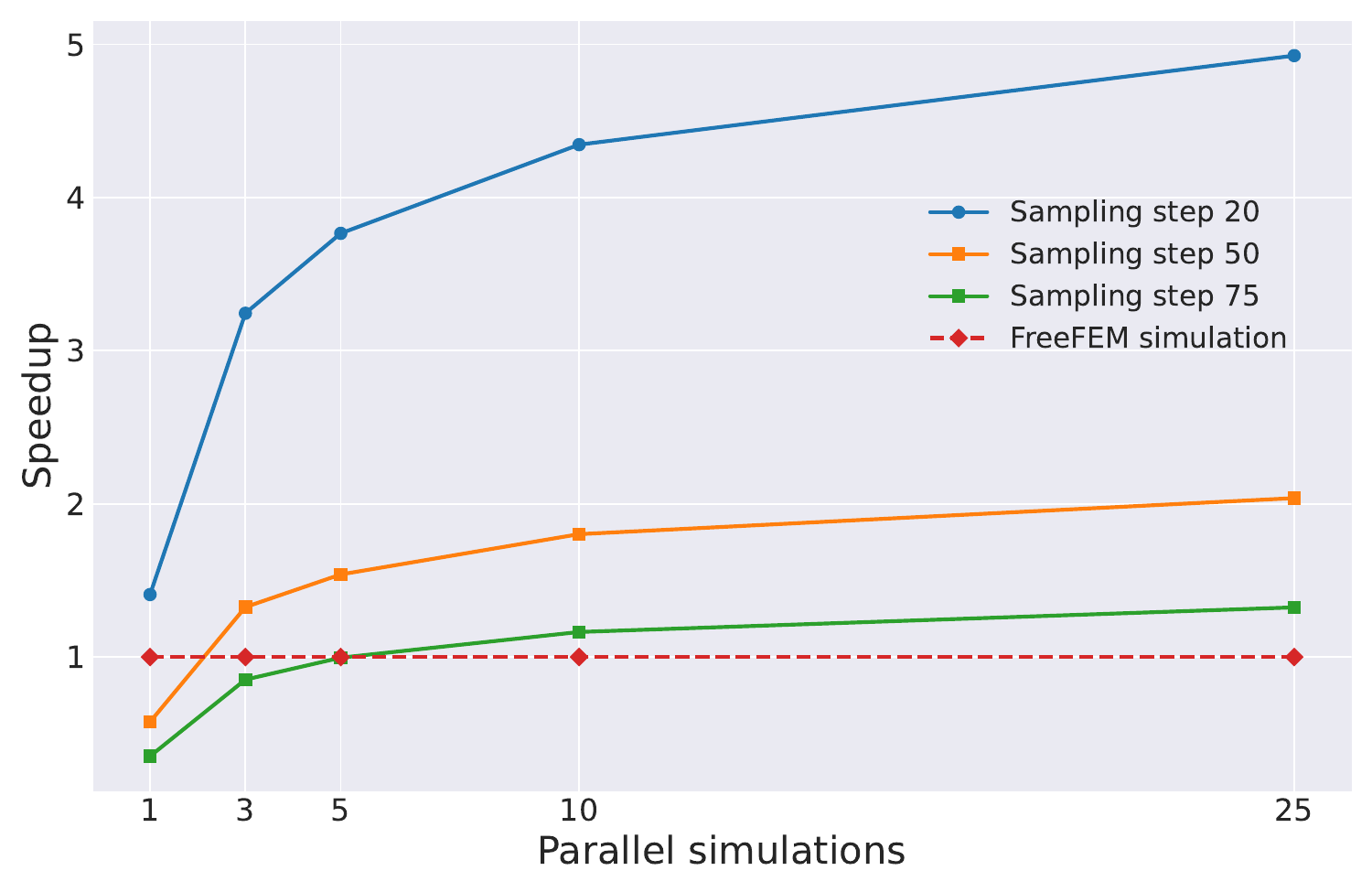}
    \caption{Speedup obtained using parallel calculation with our baseline compared to sequential simulations with FreeFEM.}
    \label{fig:speedup}
\end{figure}

\subsection{Discussion}
The results of our study highlight the potential of deep learning methods for accelerating material simulations, particularly for pressure field estimation using the Helmholtz equation. Traditional numerical solvers, such as FreeFEM, while highly accurate, require high computational costs and have limited scalability, as each simulation must be performed sequentially. In contrast, deep learning models allow for parallelized inference, leveraging batch processing and multi-GPU acceleration to significantly reduce computation time.

Generating high-quality, labeled data is a fundamental step toward making this kind of methods a viable alternative to traditional solvers. By providing a dataset of 31,000 simulated materials, each with its corresponding pressure field solution, we enable the training and evaluation of data-driven models capable of learning complex physical relationships.

\begin{table}[!t]
    \centering
    \caption{Objective evaluation for different sampling steps}
    \resizebox{0.75\linewidth}{!}{
    \begin{tabular}{c|c|c|c}
    \toprule
        \textbf{Sampling steps} & \textbf{FID $\downarrow$} & \textbf{SSIM $\uparrow$} & \textbf{MSE $\downarrow$} \\
    \midrule
        20 & 53.31 & 0.657 & \textbf{0.1156} \\
        50 & \textbf{40.93} & 0.662 & 0.1290 \\
        75 & 43.62 & \textbf{0.669} & 0.1163 \\
    \bottomrule
    
    \end{tabular}}
    \label{tab:metrics}
\end{table}

To assess the efficiency of our approach, we evaluated the performance of our model using the DDIM sampler with different batch sizes (1, 3, 5, 10, and 25) and various sampling steps (20, 50, 75, and 100). Our analysis demonstrates that the generated pressure fields maintain high fidelity to the ground truth, as indicated by the objective evaluation metrics (\textbf{MSE}, \textbf{FID}, and \textbf{SSIM}) reported in Table~\ref{tab:metrics}. We observed that increasing the batch size and optimizing the number of sampling steps allows for a significant reduction in inference time while maintaining competitive accuracy, as shown in Figure~\ref{fig:speedup}.

By parallelizing inference and tuning sampling parameters, our approach enables rapid and scalable material simulations, making it a viable alternative for large-scale acoustic material analysis. Furthermore, our dataset can be used as a benchmark for future research.

\section{Conclusions}
\label{sec:con}
In this work, we introduced the HA30K dataset, a dataset for deep learning-based acoustic material simulation, specifically addressing pressure field estimation via the Helmholtz equation. By leveraging deep generative models, we demonstrated the advantages of using data-driven approaches instead of traditional numerical solvers to significantly accelerate simulation times through parallelized inference. Our experimental results confirm the high fidelity of the generated pressure fields compared to ground truth simulations, while also showcasing the efficiency of batch processing and optimized sampling strategies.  

Our findings highlight the potential of deep learning for material simulation, paving the way for more scalable and efficient acoustic analysis methods. Furthermore, the dataset we provide establishes a new base for future research in data-driven material design and inverse modeling, fostering advancements in physics-informed deep learning applications.

\section*{Acknowledgment}

This work was supported by the European Union under the Italian National Recovery and Resilience Plan (NRRP) of NextGenerationEU, partnership on “National Centre for HPC, Big Data and Quantum Computing” (CN00000013 - Spoke 6: Multiscale Modelling \& Engineering Applications).

\bibliography{refs}
\bibliographystyle{ieeetr}

\vspace{12pt}

\end{document}